# MR fingerprinting Deep RecOnstruction NEtwork (DRONE)

*Ouri Cohen[1,2,3], Bo Zhu[1,2,3], Matthew S. Rosen[1,2,3]*

[1]Athinoula A. Martinos Center for Biomedical Imaging

Massachusetts General Hospital

149 13th Street, Charlestown, MA 02129 USA

[2]Department of Radiology, Harvard Medical School, Boston, MA 02115 USA

[3]Department of Physics, Harvard University, Cambridge, MA 02138 USA

**ABSTRACT**

**PURPOSE:** Demonstrate a novel fast method for reconstruction of multi-dimensional MR Fingerprinting (MRF) data using Deep Learning methods.

**METHODS:** A neural network (NN) is defined using the TensorFlow framework and trained on simulated MRF data computed using the Bloch equations. The accuracy of the NN reconstruction of noisy data is compared to conventional MRF template matching as a function of training data size, and quantified in a both simulated numerical brain phantom data and acquired data from the ISMRM/NIST phantom. The utility of the method is demonstrated in a healthy subject *in vivo* at 1.5 T.

**RESULTS:** Network training required 10 minutes and once trained, data reconstruction required approximately 10 ms. Reconstruction of simulated brain data using the NN resulted in a root-mean-square error (RMSE) of 3.5 ms for $T_1$ and 7.8 ms for $T_2$. The RMSE for the NN trained on sparse dictionaries was approximately 6 fold lower for $T_1$ and 2 fold lower for $T_2$ than conventional MRF dot-product dictionary matching on the same dictionaries. Phantom measurements yielded good agreement ($R^2$=0.99) between the $T_1$ and $T_2$ estimated by the NN and reference values from the ISMRM/NIST phantom.

**CONCLUSION:** Reconstruction of MRF data with a NN is accurate, 300 fold faster and more robust to noise and undersampling than conventional MRF dictionary matching.

## Introduction

Magnetic Resonance Fingerprinting (MRF) (1) is an acquisition strategy that uses a variable schedule of RF excitations and delays to induce differential signal evolution in tissue of differing types. Quantitative tissue parameter maps are then obtained by matching the acquired signal to a pre-computed dictionary consisting of the time-evolution of the magnetization of all possible tissue parameter values. Multiple quantitative tissue parameter maps can be simultaneously obtained from a single experiment, significantly reducing the total scan time.

To avoid errors in the reconstructed tissue maps, the reconstruction dictionary is typically computed with fine granularity over the entire range of possible tissue values. Dictionary size, however, grows exponentially as the number of tissue parameters (i.e the dictionary dimension) is increased which can quickly result in prohibitively large dictionaries that require extensive computational resources to process (2). This increased memory, storage and computational burden is a limiting factor for clinical adoption of MRF methods and is particularly pernicious in innovative high dimensional applications of MRF (3,4). Reducing the dictionary density is a poor solution for this problem since it limits the *a priori* accuracy of the reconstruction even before experimental factors are accounted for.

Existing methods address aspects of this problem but important challenges remain. Dictionary compression (5,6) uses the compressibility of the fingerprints to reduce the dimensionality of the dictionary leading to faster post-processing. However, to create the compressed dictionary the full fine-grained dictionary must first be generated prior to decomposition with the singular value decomposition (SVD), itself a computationally expensive operation. Recent work by Yang et al (7) used a randomized SVD with an iterative polynomial fit to reduce the memory requirements though at the cost of increased processing time. Additionally, when optimized acquisition schedules are used (8,9), the compressibility of the dictionary may be significantly reduced rendering these methods less effective. Optimizing the acquisition schedule can indeed reduce the number of image frames needed for accurate reconstruction but the reduction is smaller than the exponential growth of the dictionary with increasing dimensions.

In recent years, the availability of inexpensive graphical processing units (GPU) has led to significant advances in neural networks (NN) and Deep Learning (DL) algorithms used to train these networks (10). Mathematical work in NN theory has shown that any Borel measurable function can be represented by a NN with a finite number of neurons (11) which can therefore offer a compact representation of complicated functions. In this paper we exploit this property and describe a novel method that reframes the MRF reconstruction problem as learning an optimal function that maps the acquired signal magnitudes to their corresponding tissue parameter values, trained on a sparse set of dictionary entries (12).  The trained neural network reconstruction function is remarkably compact (~20 times smaller than typical MRF dictionaries) and reconstruction is nearly instantaneous (~300-fold faster than conventional dictionary matching techniques) due to its rapid feedforward processing. We validate our method by numerical simulations and phantom experiments and demonstrate its utility in the brain of a healthy subject scanned at 1.5 T.

## Methods

### Neural Network

A four layer fully-connected neural network composed of input and output layers and two hidden layers was defined using the TensorFlow framework (13) as shown in Fig. 1. The input layer consisted of 25 nodes to correspond to the 25-point trajectory of magnitude images acquired with our optimized echo-planar imaging (EPI) MRF sequence (8). Complex-valued images can also be processed by this network topology by splitting the real and imaginary channels, or through the use of a complex-valued network (14). In this proof-of-concept study only $T_1$ and $T_2$ were considered so the output layer consisted of two nodes; reconstruction of additional tissue maps would require a larger output layer. Each of the two hidden layers had 300 nodes. The network was trained by the ADAM stochastic gradient descent algorithm (15) with the learning rate set to 0.001 and the loss function (cost) defined as the mean square error:

$$LF = \frac{1}{n}\sum_{k=1}^{n}(P_{training}^{k} - P_{recon}^{k})^2,$$
(1)

where $k$ ranges over the $n$ training samples and $P$ is the training or reconstructed tissue parameter of interest ($T_1$ or $T_2$ in this study). Two different activation functions were defined. A hyperbolic tangent (tanh) function was used for the hidden layers with a sigmoid function used for the output layer. In total, the NN required storage of 300×300=90,000 coefficients.

### Pulse Sequence

All experiments in this study used a modified gradient-echo EPI MRF pulse sequence whose flip angles (FA) and repetition times (TR) were set according to an optimized measurement schedule, as previously described (8,16,17). We note that the MRF reconstruction with our approach is not limited to this a specific pulse sequence, as the NN can be trained to accommodate any MRF pulse sequence and MRF sequences that require a large number of acquisitions (1,18–21) would simply require a larger input layer to accommodate the higher acquisition count.

### Network Training

A dictionary of ~79,900 entries consisting of $T_1$ in the range 1:10:5000 ms (min:increment:max) and $T_2$ values in the range (1:10:2000) ms was defined, excluding entries where $T_1 < T_2$. The magnetization due to each ($T_1$, $T_2$) pair was calculated by Bloch equation simulation using the Extended Phase Graph formalism (18,22). Gaussian noise with 2% standard deviation and zero mean was added to the training dictionary to promote robust learning, as previously exhibited with denoising autoencoders (23). This dictionary was used to train the network for 1000 epochs to ensure convergence, requiring approximately 10 mins on an Nvidia K80 GPU with 2 GB of memory. The efficacy of training was validated by using the training dictionary as input to the network and comparing the resultant $T_1$ and $T_2$ values to the true values. The size of this dictionary required storage of 25×79,900=2 million floating point coefficients or 15 MB.

**Numerical Simulations**

The performance of the network in reconstruction of realistic brain $T_1$ and $T_2$ values was assessed using the Brainweb digital brain phantom (24). An MRF acquisition was simulated as described above, and the resulting signal used as an input to the network. The reconstructed tissue maps were used to calculate the root-mean-square error (RMSE) in the reconstructed $T_1$ and $T_2$ maps.

*Training dictionary density vs reconstruction error*

The effect of the training dictionary density on the resulting reconstruction error was measured by sub-sampling the initial dictionary variously from 2 to 60 fold. The network was then trained with each sub-sampled dictionary and used to reconstruct the initial, fully sampled dictionary whose entries were corrupted by zero mean Gaussian noise with 0.5% standard deviation. This was repeated 10 times for each undersampling factor to allow statistical calculations. The NN reconstruction was also compared to a conventional MRF dictionary matching reconstruction (1) by matching the initial, fully sampled dictionary to each sub-sampled dictionary and calculating the resultant error. The mean and standard deviation of the RMSE of the reconstructed $T_1$ and $T_2$ maps of each reconstruction method was then calculated as a function of the dictionary sub-sampling factor.

**MRI**

All experiments were conducted on a 1.5 T whole-body scanner (Avanto, Siemens Healthcare, Erlangen, Germany). The manufacturer's body coil was used for transmit operation and a 32-channel head coil array used for receive operation. The TI/TE/BW was set to 19/23/2009 Hz/pixel. The slice thickness was set to 5 mm and the in-plane resolution set to $2\times2$ mm$^2$ with a matrix of $128\times128$ and an acceleration factor of 2 for a total scan time of ~3 seconds for the 25 frames acquired with the optimized schedule. The images were reconstructed online using the GRAPPA (25) method.

**Phantom**

The accuracy and precision of the NN reconstruction was assessed using the International Society for Magnetic Resonance in Medicine (ISMRM)/National Institute of Standards and Technology (NIST) multi-compartment phantom with calibrated $T_1$ and $T_2$ values similar to those of the human brain (26). The phantom was scanned with the MRF EPI sequence and the images were reconstructed with the NN defined above. The resulting $T_1/T_2$ maps were compared to the true phantom values which were characterized by NIST and calculated using gold-standard NMR spectroscopy IR and CPMG sequences (26,27).

***In Vivo* Human**

A healthy 25-year-old female subject was recruited for this study and provided informed consent prior to the experiment in accordance with our Institution Human Research Committee. Following reconstruction of the acquired data with the proposed NN, three regions-of-interest (ROIs) were defined corresponding to grey matter, white matter and cerebrospinal fluid (CSF).

The mean±standard deviation of the $T_1/T_2$ values within those ROIs was calculated and compared to values from the literature.

## Results

### Network Training

The evolution of the training cost as a function of the epoch number is plotted on a log scale in Fig. 2 and illustrates the convergence of the network training. The reconstructed training data is shown in Fig. 3 in comparison to the true training values. Excellent agreement was obtained between the true and reconstructed $T_1$ and $T_2$ values yielding a correlation coefficient of $R^2$=0.99 for both $T_1$ and $T_2$ with a negligible bias of 10 ms for $T_1$ and 2.2 ms for $T_2$ and an RMSE of 9.3 ms for $T_1$ and 8.8 ms for $T_2$. Short $T_1$ and $T_2$ values (of the order of the echo time of the EPI sequence) and very long $T_2$ values (> 1000 ms) showed increased deviation from the true values, possibly due to the increased difficulty of modeling the very rapid or very slowly varying temporal signal dynamics in these tissues.

### Numerical Simulation

The true and reconstructed $T_1$ and $T_2$ maps of the numerical brain phantom are shown in Fig. 4 along with the associated error map calculated as |True-Reconstructed|. The RMSE for each map was 3.5 and 7.8 ms for $T_1$ and $T_2$ respectively.

#### *Training dictionary density vs reconstruction error*

The mean $T_1$ and $T_2$ RMSEs across the ten repetitions are shown as a function of the dictionary undersampling factor in Fig. 5. The RMSE of the NN reconstruction grew with increasing undersampling, as expected, but was approximately 5-7 fold smaller for $T_1$ and 2 fold smaller for $T_2$, for all undersampling factors, than the relatively constant RMSE of the conventional dictionary matching. The growth in the NN reconstruction error was well fit ($R^2$ = 0.94 for $T_1$ and $R^2$=0.89 for $T_2$) by a linear function.

**Phantom**

The NN reconstruction accuracy was evaluated by estimating $T_1$ and $T_2$ values in the well-characterized calibrated ISMRM/NIST phantom. The measured $T_1$ and $T_2$ values were derived from the mean $T_1$ and $T_2$ values estimated within each compartment. The two phantom compartments with the shortest $T_2$ (<30 ms) showed significant errors and were not included in the comparison. The estimated $T_1$ and $T_2$ values from the remaining compartments (Fig. 6) showed good agreement to the true phantom values ($R^2$=0.99) and a minimal estimation bias of 5.5 ms for $T_1$. The bigger $T_2$ bias was largely caused by the long (>1200 ms) $T_2$ compartments where the temporal signal dynamics are more difficult to model, as mentioned in the *Network Training* section above. The calculated $T_1$ and $T_2$ RMSEs were 72 and 88 ms respectively. The error in $T_2$ was significantly affected by four phantom compartments with short (<30 ms) or long (> 1200 ms) $T_2$s. When those were excluded from the analysis, the $T_2$ bias was reduced to 21 ms and the RMSE to 31 ms. The reconstruction of the 128×128 $T_1$ and $T_2$ maps required ~10 ms with the NN which was ~300 fold faster than the ~3s required with conventional dictionary matching using a 79900 entries dictionary.

***In Vivo* Human Brain**

The $T_1$ and $T_2$ maps reconstructed by the NN are shown in Fig. 7 along with the ROIs chosen. The $T_1$ and $T_2$ mean±standard deviation for each tissue compartment are shown in Table 1 and are similar to values obtained from the literature (28).

## Discussion

MRF enables quantitative tissue mapping in a short acquisition time at the cost of increased complexity in the reconstruction. While compute resources are typically cheaper and more accessible than scanner time, the large dictionaries required for MRF applications can overburden even the most advanced hardware. In this work, a four layer NN capable of modeling the time-dependent Bloch equations used in MRF sequences was demonstrated. Unlike conventional dictionary matching where the acquired signals can only be matched to the discrete entries computed in the dictionary, the proposed method relies on the functional representation within the NN that yields continuous-valued parameter outputs. Furthermore, our training process results in a signal-to-parameter mapping that is more robust to noise than a conventional dictionary matching approach because the mapping is forced to be expressed in low-dimensional space and is thus insensitive to small corruptive input perturbations. The results shown in Fig. 5 indicate that the NN reconstruction error grew linearly with the dictionary sub-sampling factor. Conventional dictionary matching does not learn a functional mapping, relying instead on the similarity between the normalized measured data and the corresponding normalized dictionary entry. An unfortunate side-effect of the normalization is that noisy signals, arising from tissues with short $T_2$ for example, are amplified and then matched to some dictionary entry leading to increased reconstruction errors. In contrast, the network was trained on noisy signals and consequently yielded smaller error for both $T_1$ and $T_2$ than dictionary matching of the same noisy data.

The compact size of the NN solves many of the problems inherent to conventional dictionary matching. Specifically, the NN required merely ~5% of the storage and memory needed for storing even the small training dictionary used - larger dictionaries would reduce this fraction further. Because of its feedforward structure, reconstruction with the network was 300-fold faster than conventional dot product dictionary matching. Accelerated matching techniques such as that reported by Cauley et al (5) still necessitated 2 seconds for reconstruction which was 200× longer than what was required with the NN reconstruction. While the dictionary used in that study was larger than the one used in this work, the number of entries in the NN training dictionary has no effect on the final reconstruction time once the network is fully trained. Because the network topology is fixed, additional training entries simply modify the weights of the network but do not increase the reconstruction time.

The quantity and quality of the training data are usually the key factors in determining the success of deep learning networks for a given application. Large, high quality clinical datasets may be difficult to obtain and expensive to generate. Training the network on simulated dictionary data eliminates this concern and permits generating arbitrarily large training sets. Our results (Fig. 6) show that networks trained on simulated data can accurately reconstruct measured data despite the presence of inevitable noise and other sources of errors in the measurements.

This work represents an initial proof-of-concept for MRF reconstruction by a NN system and can be further optimized to further improve the results. For instance, although the training time for the network used in this study was small (~10 minutes), the rapid convergence of the training (Fig. 2) indicates that training time can be further shortened by reducing the number of epochs used. The size of the network and the small number of images used with the optimized MRF schedule contributed to the short training time. Alternative MRF sequences that require a greater number of acquisitions (10-100 fold higher) will likely require a longer training time as will simultaneous reconstruction of additional tissue parameters. Although the architecture of our network (300×300 fully-connected hidden layers) theoretically allows up to $300^2$ degrees of freedom, improving the accuracy of the network reconstruction or reconstruction of additional parameters such as $B_1$ and $B_0$ may require deeper networks. The sigmoid and tanh activation functions used in this study are a common choice for NN training (29) but skew the accuracy of the network towards the middle of the training dictionary range (Fig. 2) where the gradient is largest and the back-propagation algorithm is thus most effective. A comparison of the effectiveness of different activation functions (softmax, ReLu…) is beyond the scope of this work and will be examined in future studies.

## Conclusion

We have demonstrated the feasibility of using deep learning networks for reconstruction of MRF data. The proposed approach yields fast and accurate reconstruction with a limited storage requirement despite training on sparse dictionaries and can therefore resolve the technical issues inherent to the exponential growth of multi-dimensional dictionaries.

## Acknowledgments

B.Z. is supported by NIH National Institute for Biomedical Imaging and Bioengineering grant F32-EB022390. This work was also supported in part by the MGH/HST Athinoula A. Martinos Center for Biomedical Imaging and the Center for Machine Learning at Martinos.

## Table 1

| Compartment | T$_1$ (ms) | T$_2$ (ms) |
|---|---|---|
| White Matter | 663±30 | 83±7.9 |
| Grey Matter | 1110±59 | 96±13 |
| Cerebrospinal Fluid | 3799±422 | 870±191 |

**Table 1**: Mean±standard deviation T$_1$s and T$_2$s values of the ROIs selected.

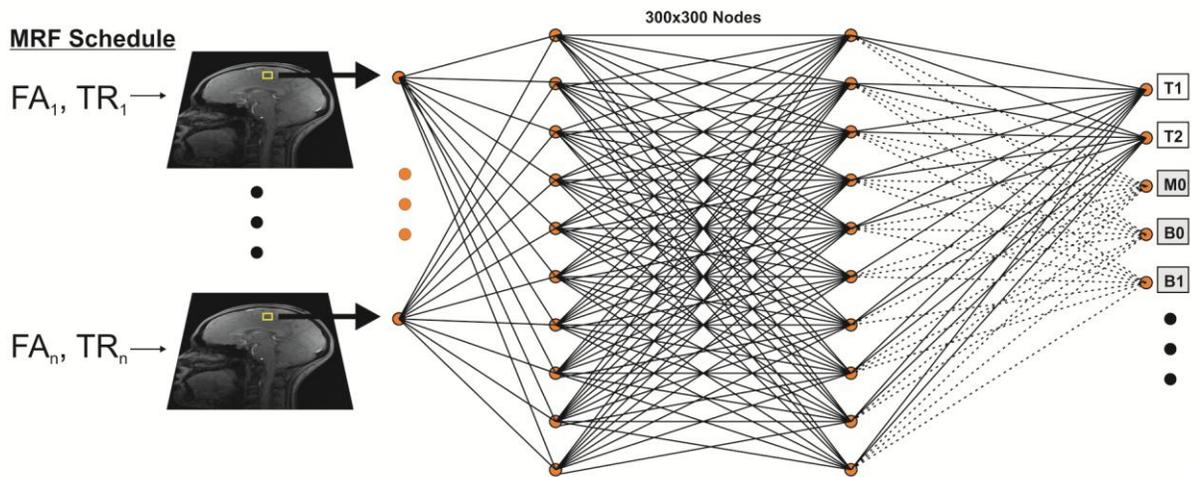

**Fig. 1.** Schematic of the reconstruction approach used in this study. MRI data acquired with the

optimized MRF EPI sequence is fed voxelwise to a four layer neural network containing

two 300×300 hidden layers. The network is trained by a dictionary generated using Bloch

equation simulations with the tanh and sigmoid functions used as activation functions of

the first and last hidden layers respectively. The network then outputs the underlying

tissue parameters $T_1$ and $T_2$. Additional tissue parameters including $M_0$, $B_0$, $B_1$ etc…

(gray boxes) can similarly be obtained by training the network with a suitable dictionary.

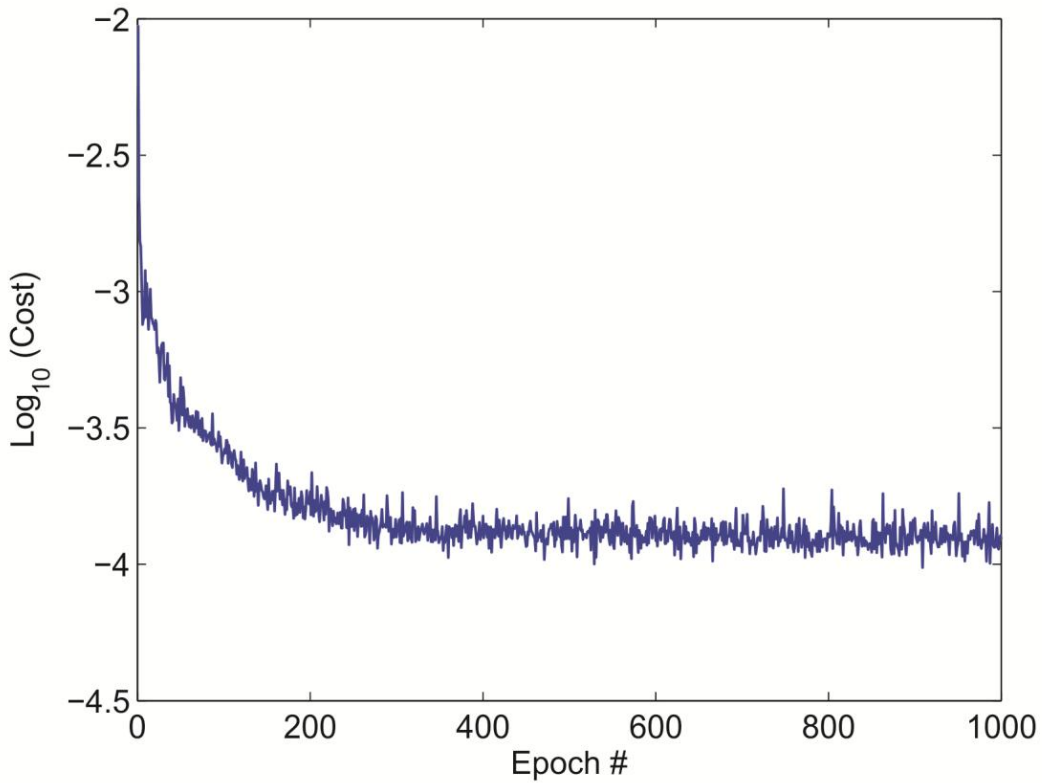

**Fig. 2.** Evolution of the training cost as a function of the epoch number shown on a log scale for

improved visualization. The 1000 epochs used in the training required approximately 10

minutes on an Nvidia K80 GPU though the majority of the cost reduction occurred within the initial 400 epochs implying that training time may be reduced without adversely impacting the results.

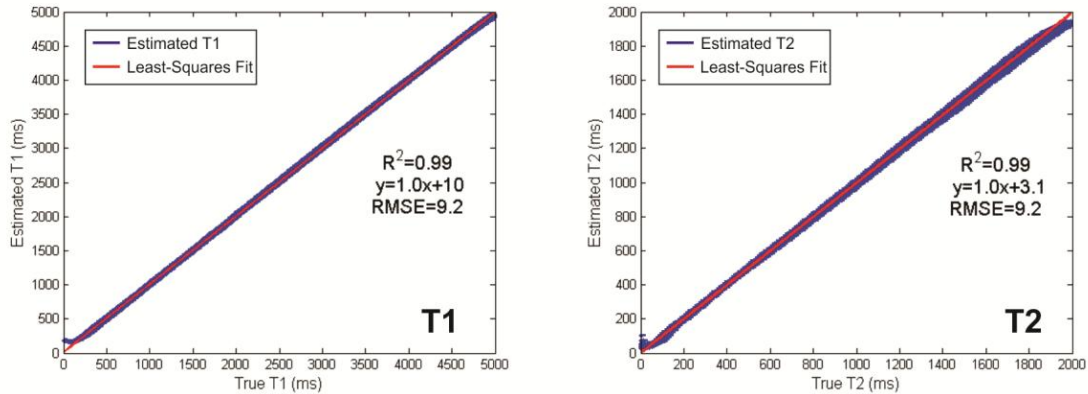

**Fig. 3.** Shown is a comparison between the training (true) $T_1$ and $T_2$ and those reconstructed by a network trained on those values. The red line indicates the least-squares fit curve. The reconstructed $T_1$ and $T_2$ values showed excellent agreement ($R^2$=0.99) with the true values with a minimal bias in $T_1$ and $T_2$ of 10 and 3.1 ms respectively, validating the feasibility of the proposed approach. Short $T_1$ and $T_2$ values as well as very long $T_2$ values showed increased deviation from the true values due to increased difficulty in modeling the very rapid or very slowly varying temporal signal dynamics in these tissues.

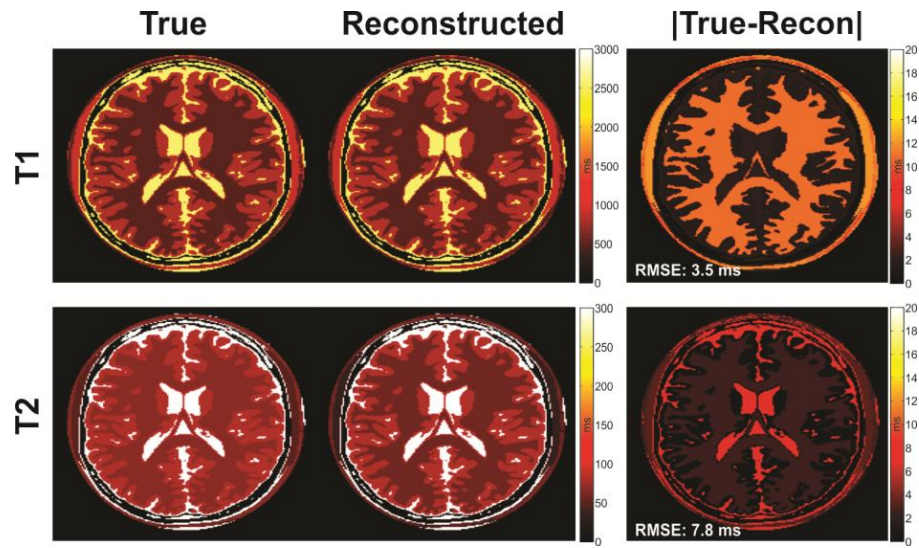

**Fig. 4.** True and reconstructed $T_1/T_2$ images from the numerical brain phantom shown on a common ms scale and the associated error map. Note the close agreement between the reconstructed and true maps. The $T_1$ and $T_2$ RMSEs of 3.5 and 7.8 ms respectively are shown inset in white in the error map.

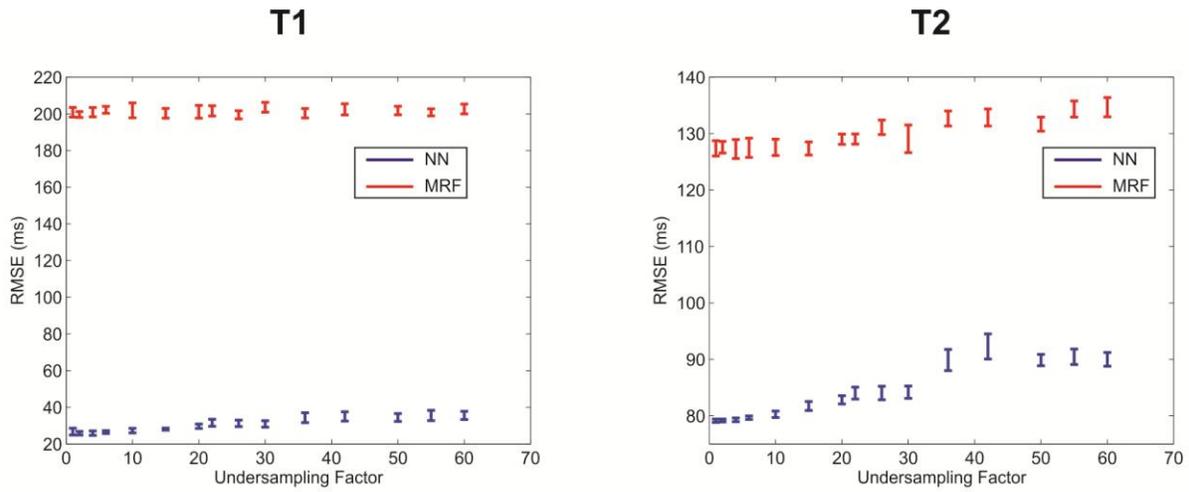

**Fig. 5**. Mean (circle) and standard deviation (whiskers) of the reconstruction RMSE of noisy

data using conventional dictionary matching (red) or the proposed NN (blue) for

undersampled dictionaries. The full dictionary was undersampled by the undersampling

factors shown and was used to either directly match the noisy data or to train a NN which

was then used to reconstruct the noisy data. Note the lower error for the NN

reconstruction despite increased undersampling.

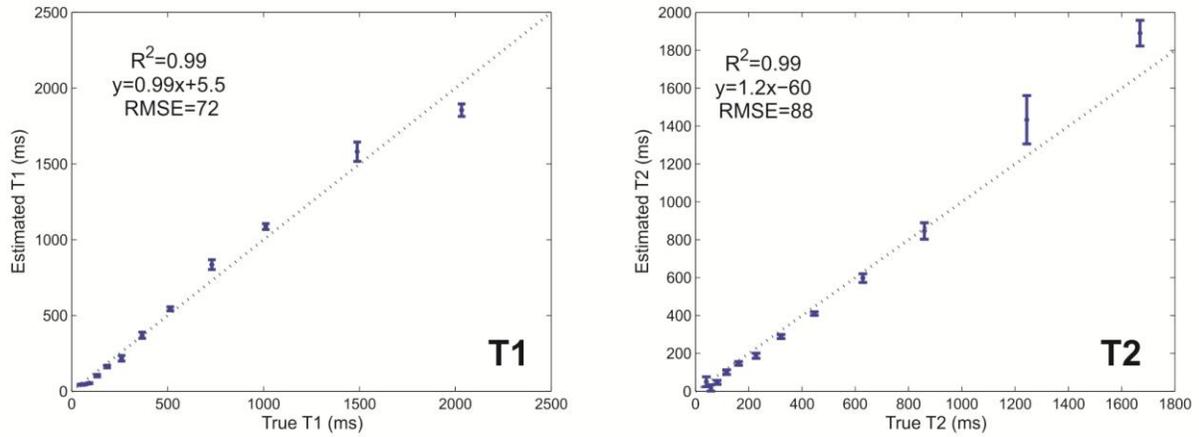

**Fig. 6**. T$_1$ and T$_2$ estimation accuracy evaluated in the ISMRM/NIST phantom. Shown is a comparison between the true and measured compartment T$_1$ and T$_2$ values for data acquired with the optimized MRF EPI sequence and reconstructed by the NN. The dashed line is the identity line. The error bars represent the standard deviation of the measured T$_1$ and T$_2$ values within each compartment.

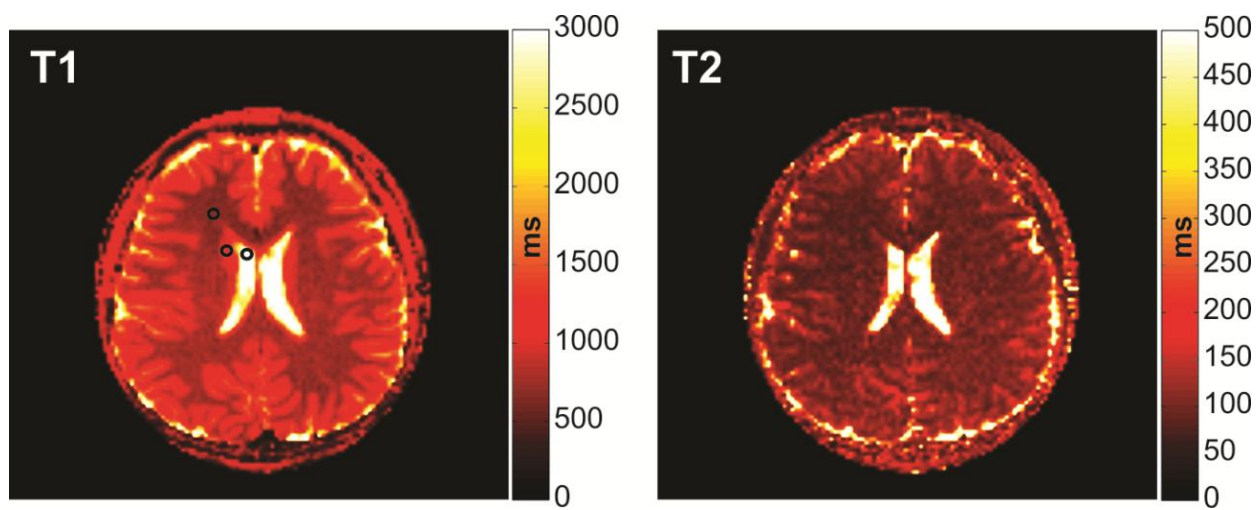

**Fig. 7.** In vivo quantitative $T_1$ and $T_2$ maps from the brain of a healthy subject reconstructed with the proposed NN. The black circles indicate the locations of the grey matter, white matter and CSF regions used to calculate the mean $T_1$ and $T_2$ values shown in Table 1.